# DOMAIN ENGINEERING FOR APPLIED MONOCULAR RECONSTRUCTION OF PARAMETRIC FACES

Igor Borovikov, Karine Levonyan, Jon Rein, Pawel Wrotek and Nitish Victor

Electronic Arts, Redwood City, CA, USA

## ABSTRACT

*Many modern online 3D applications and videogames rely on parametric models of human faces for creating believable avatars. However, manually reproducing someone's facial likeness with a parametric model is difficult and time-consuming. Machine Learning solution for that task is highly desirable but is also challenging. The paper proposes a novel approach to the so-called Face-to-Parameters problem (F2P for short), aiming to reconstruct a parametric face from a single image. The proposed method utilizes synthetic data, domain decomposition, and domain adaptation for addressing multifaceted challenges in solving the F2P. The open-sourced codebase illustrates our key observations and provides means for quantitative evaluation. The presented approach proves practical in an industrial application; it improves accuracy and allows for more efficient models training. The techniques have the potential to extend to other types of parametric models.*

## 1. INTRODUCTION

Modern virtual 3D environments strive to deliver a life-like representation of human facial likeness under a limited computational budget. The need for believable virtual characters emerges in production art tools and user-facing applications. In particular, customizable video games or 3D Virtual Reality (VR) characters generated from user-provided photographs are in high demand (e.g., [1,2,3]). While tools, inputs, and computational constraints may differ in user-facing and internal pipelines, the challenges remain mostly shared and may benefit from similar solutions. The paper places the monocular reconstruction problem in a general parametric context and shows a promising approach that may efficiently work in many practical applications. The following subsection introduces some necessary terminology.

### 1.1. 3DMM vs. Parametric Model

The paper's context requires emphasizing some crucial distinctions in the common avatar creation frameworks. Namely, Computer Graphics (CG) and Computer Vision (CV) rely on two fundamentally distinct methods for creating human faces. One approach utilizes a fully Morphable 3D Mesh (3DMM) to adjust individual vertices to produce a desired shape [4,5]. A radically different approach is parametric [7,8]. A parametric model abstracts from the vertices and, in that sense, is more general. It relies on a pre-fixed collection of hand-authored construction elements (frequently called "blendshapes" [6]). The blendshapes contribute to the target shape with the weights defined by the input parameters, e.g., the distance between eyes, size of the mouth or nose, and alike. All the characters within an application use only these elements for constructing head shapes.

A direct comparison of 3DMM vs. parametric frameworks for ML applications is rarely insightful within an industrial context: the engineering, production, and art style decisions take precedence in designing a concrete application. Hence, it is not an apple-to-apple comparison. On the one hand, models can be hand-crafted with highly detailed 3DMMs in the film industry. Conversely, expensive authoring, tight development timeframe, limited memory, or bandwidth benefit from the time, data, and cost reduction when relying on a modest set of predefined assets. Such assets can be preloaded to the customer hardware and allow for compact encoding of the models via a relatively small parameter set in interactive environments.

The paper focuses on *parametric* models suitable for interactive user-facing software and leaves out 3DMM-based systems for a different exploration.

## 1.2. Face-to-Parameters with a Heterogeneous Target Space

In our formulation, the Face-to-Parameters (F2P) problem aims to reproduce the face on a single input image in the best manner possible by optimally selecting two distinct kinds of model parameters: continuous and discrete. The presence of discrete construction elements makes our target space heterogeneous. Also, it introduces the combinatorial aspect, which we discuss in later sections. The combinatorial considerations contributed significantly to the motivation for our approach.

Continuous parameters, as we already mentioned, are the weights of blendshapes. Blendshapes are common in modeling complex static or articulated objects, like human bodies and faces (see blendshape methods review in [6]). A user-facing application or an in-house modeling tool has a fixed set of blendshapes used across all the characters. The blendshapes aim to represent all anticipated target geometries in the best manner possible while complying with design limitations. All blendshapes contribute to the resulting geometry simultaneously and are not mutually exclusive. A basis in a vector space is a good metaphor for them. Some examples of a blendshapes effect on the output are the length of the nose, the shape of the nose (straight, curvy, down, up, narrow, or wide), mouth location relative to the nose and the chin, the shape of the mouth (plum, thin, curved up or down), how prominent is the chin, and alike.

The discrete elements represent gross deformations of an underlying mesh. They are also fixed but mutually exclusive within their region of the target geometry. In our proprietary application, they are called sculpts. To enhance the effect of a sculpt, an artist may paint details for the features too difficult to represent with the mesh. Adding a texture to the texture compositing stack with a sculpt usually makes it necessary for sculpts to be exclusive. An utterly distinct asset, say, a carnival mask, may also define a sculpt. Other less extreme examples include a specific nose shape that is not reachable with the available blendshapes. E.g., a broken nose, a Sci-Fi alien nose, and so on.

In the ML context, fitting the continuous parameters is a regression problem and fitting the discrete elements is a classification problem. In this paper, the F2P problem combines both. In our proprietary application, the dimensionality of the continuous parameters space is $\sim$100, and the discrete elements count exceeds 300. By design, the parameters come partitioned into facial regions (e.g., nose, mouth, and alike). That leads to the combinatorial complexity in the order of $10^{11}$. For comparison, the FLAME model [7] has no discrete elements and exposes ~300 continuous parameters, with ~100 devoted to articulating the character. We consider only neutral faces and exclude facial expressions from consideration in this paper. An open-source software

Makehuman [8], which we use for reproducible quantitative evaluation of this paper, exposes over 100 continuous parameters and only a handful of discrete ones. The smaller parametric space makes it perfect for illustration purposes. Architecturally, Makehuman is a close counterpart of the proprietary software we use in this work. That makes it more convenient to open-source our experimental code using Makehuman vs. systems like FLAME. Figure 1 illustrates the data structure for the head parameters used by Makehuman.

```
…
modifier head/head-age-decr|incr 0.068249
modifier head/head-angle-in|out -0.070203
…
modifier head/head-oval 0.255893
…
modifier head/head-scale-depth-decr|incr 0.112490
modifier head/head-scale-horiz-decr|incr 0.103516
…
modifier forehead/forehead-trans-backward|forward 0.069743
modifier forehead/forehead-scale-vert-decr|incr -0.046971
…
modifier eyebrows/eyebrows-angle-down|up -0.087145
modifier eyebrows/eyebrows-trans-down|up -0.029648
…
modifier eyes/r-eye-bag-in|out -0.039712
modifier eyes/r-eye-bag-height-decr|incr 0.082509
…
eyelashes eyelashes02 04a0718e-aaa4-4480-a013-ad51703bef6b
eyebrows eyebrow004 eb028b6d-3ff8-40c7-a2ea-4d9aa808b38d
…
```

Figure 1. Snippets of an mhm file illustrating Makehuman data structure used for its parametric model. Parameter values are normalized to -1…1 range. Two selected illustration discrete elements are at the bottom and are identified by their guid in the Makehuman library of assets.

For the clarity of this presentation, we omit color palette elements, hairstyles, makeup, tattoo, facial hair, accessories, etc. - anything that does not directly change the geometry of a character's face. We exclude the geometry of the ears and neck as they are frequently hard to determine in the imagery. Also, facial texture reconstruction and mapping on the 3D model [9,10] are (mostly) outside of our discussion. Facial expressions, pose, lighting, and camera parameters are excluded, too, even though, in principle, they can be included in the ML pipeline we aim to develop.

At the high level, the F2P problem formulation here is similar to that one in [3] yet relies on a different approach that we outline next.

## 1.3. The Approach Outline

For the reasons we explain in the following sections, we cast the heterogeneous F2P problem as a classic supervised learning problem. That requires abundant training data, i.e., facial images with the corresponding parameters. Identifying the source of the training data requires careful navigation around privacy and licensing aspects. Most of the available human face datasets, e.g., a popular CelebA dataset [11], exclude commercial applications. With growing concerns around privacy, many previously accessible facial datasets are no longer available. The most direct way to work around these challenges is to use synthetic data. The generation of synthetic data is often straightforward and can produce practically unlimited amounts of it reasonably quickly with a wealth of the associated metadata. Full control over training data also facilitates the normalization of training images: i.e., rotation and scaling to place the centers of eyes into predefined locations on the image to match those during inference.

The synthetic nature of the data creates new opportunities not readily available when working with real-world data. The parameters of the human face model naturally map to different facial regions. In parallel to that grouping, the rendering pipeline can include automated generation of the corresponding semantic segmentation of the synthetic images. Such natural separation leads to domain decomposition: the ML pipeline can become a hierarchical ensemble of models dealing with the general structure of the face and local models that control separate regions. The ensemble allows smaller models, which are less prone to overfitting and can train and execute in parallel utilizing data and model parallelism (e.g., [12] reviews conceptually similar works).

While synthetic data facilitates the decomposition and training of the related ML models, it also presents a challenge due to the inevitable domain gap between synthetic and real-world imagery (a brief introduction to domain adaptation in [13]). Imagery such as selfies may be only one of the possible input types for the F2P models. Examples of other possible target domains include sketches, fine art portraits, faces from comics, anime, and more. Previously unseen domains may degrade a naive F2P ML model's accuracy or render it unusable. That makes the domain gap issue broad and even more critical, suggesting that domain adaptation must be an integral part of the system we build. At the same time, the domain adaptation functionality must be modular and easily replaceable to accommodate different future applications.

We utilize style transfer for domain adaptation (see the seminal paper [14]). The direction of the style transfer is from the target imagery (e.g., selfies) to the synthetic images, which could have a distinct (fixed) art style. It is "inverse" in that sense. We train GANs specific to the multiple target domains and use them as an adapter. Training such GANs is independent of the decomposition-based ensemble training. We find that the decomposition and domain adaptation enhance each other and lead to an ensemble that produces better results in solving the F2P problem than a direct approach based on a single monolithic model.

## 1.4. Contribution

The paper emphasizes the practical aspects of converting facial images into the parameters of a parametric model of a human face and proposes a novel, efficient solution. While the individual building blocks of our approach are not new (hierarchical decomposition, domain adaptation, models ensemble), their proposed combination is not found in the literature.

Concretely, we cast the F2P problem as a supervised learning problem on synthetic data, introduce a model ensemble to take advantage of the hierarchical nature of the domain, and train separate dedicated models for domain adaptation. The proposed architecture is different from the previous works (e.g., [3]) and offers several practical advantages in the industrial environment. The open-sourced code illustrates the claim that the proposed ensemble performs better than a monolithic model thanks to leveraging the structure of the application domain. We advocate the modularity of the proposed system and offer a quantitative evaluation of claims.

In Section 8, Discussion and Future Work, we place our approach in a more general context. The decomposition approach to complex systems has deep roots in the abstract theory of sets [53] and the theory of categories. It demonstrated effectiveness in application to complex mathematical models (e.g., [54]) and provides a rich methodological foundation for this paper.

In the rest of the paper, we review previous works and then focus on the main objective: building an efficient ML system solving the outlined F2P problem, which infers the target face parameters from the input images. The system we develop is implemented as proprietary industrial software. However, we address reproducibility and quantitative evaluation of the techniques with Makehuman (free open-source software) by open-sourcing parts of our experimental code as well [15].

## 2. PREVIOUS WORKS

Facial likeness reconstruction in computer vision is a vast active area of research. For a relatively recent comprehensive review of the field, we refer to [16]. Here, we highlight only several of the relevant publications. A well-established body of work aims at generating sculptable (and texturable) mesh directly from monocular input: e.g., [2,17,18,19,20,21,22,23,24,25] or video [45,46]. While learning face reconstruction approaches may vary, synthetic imagery is instrumental due to mounting concerns around privacy and the lack of suitable datasets. The synthesis can come in the form of direct exploration of the target shape with realistic images (e.g., analysis-by-synthesis in [25]), or learning-based inverse unrealistic rendering (e.g., [52]). Enhancing a straightforward CNN architecture mapping inputs to the target space with additional modules may provide advantages. E.g., in [25,26], a dedicated DNN enhances unrealistic coarse faces to bring in realistic details and shading. To deliver robust performance on real-world images, the synthetic face reconstruction or recognition training data must cover various lighting, pose, and expression conditions. Introducing prior assumptions can regularize the reconstruction problem and improve results by imposing limitations on shape, lighting, and photometric model. E.g., using Lambertian reflectance and harmonic representations of lighting facilitates solving a shape-from-shading problem in the context of face reconstruction [27]. However, nowadays, works mostly rely on abundant realistic data generation using fast-growing CG and general computational capabilities (e.g., [3,17,28,29,30,31], to mention just a few).

Parametric models may also utilize morphable meshes with a fixed topology where the parametric space explicitly encodes deltas for vertices subsets. Their reconstruction from 2D imagery achieves a very high accuracy [28] but conceptually is similar to 3DMM reconstruction. Such models differ from more specialized ones that use higher-level construction elements, e.g., blendshapes and bone-driven morphs, which are usually specific to a particular graphics engine. The models relying on high-level construction elements are somewhat less common in the literature even though they date back to the early days of CG and CV, e.g., [32]. They provide a critical advantage in data compression, which motivated one of the early works [33]. Compact data representation remains in demand for interactive graphics applications as they strive for computational efficiency. Also, as the result of compact representation, parametric models allow for easier manual authoring compared to freely deformable meshes.

Parametric models heavily depend on the underlying character modeling system's proprietary nature. That creates an obstacle to transferring the parametric techniques across applications and renders a comparison to the 3DMM-based methods irrelevant, as already emphasized. The additional difficulty comes from discrete elements of facial reconstruction. In our work, we must handle both continuous and discrete elements. The paper [3] uses a custom-trained differentiable renderer (DR), a powerful tool in the continuous domain. However, a DR could be more problematic for fitting discrete models and is still too heavy for mobile devices.

The paper [3] works around the domain gap between real and synthetic faces by introducing discriminative loss – a perceptual distance computed with embeddings of a facial recognition CNN. The discriminative part becomes an integral part of a larger model. For our purposes of building an ensemble with pluggable components, we explored several other more general GAN-based techniques. Our brief overview starts with [47] discussing various ways to perform domain adaptation. The Neural Style Transfer [14] is the pioneering work utilizing both content and style losses. The approach uses only a single image to represent the style. That appears to be rather limiting for our application. An early generative model applied for image-to-image translation between domains is pix2pix [38], which uses conditional GANs. The conditional GAN introduces two extra losses [39] that address radical differences for the domains in question. However, it requires paired image-to-image translation. CycleGAN [49] works for unpaired data. It translates the source image into the target without paired examples by introducing the (computationally intensive) consistency loss. A more recent advancement in style transfer is the well-established StyleGAN2 [34] and its later modification StyleGAN3 [50]. The StyleGAN2 architecture has multiple advantages: the scale-based hierarchy for the generator and the discriminator, pretrained models, and better stability than many other style transfer approaches. Its hierarchical nature helps to control the enhancement of the appearance of stylized images via the so-called layer-swapping. It allows for balancing features at different levels of detail using blending weights, making it an attractive candidate for our exploration.

In conclusion of this section, we note that [3] has similar objectives to ours and some superficial similarities in techniques. However, the constraints and the methodology proposed here are quite different. Our target platform excludes a differentiable renderer from the potential toolset. We exclude pixel loss from consideration since we allow various projections and lighting conditions for the input images. Finally, the proprietary codebase of [3] makes reproduction and a direct comparison with its results difficult. As such, the cited work is a valuable inspiration rather than a technical reference.

## 3. F2P AS A SUPERVISED ML PROBLEM

In this paper, we treat F2P as a classic supervised ML problem. The training and validation data in our experiments are synthetic and generated by the target software. That eliminates any licensing or privacy concerns. The open-source Makehuman [8] provides similar functionality to our proprietary software, and we use it for the quantitative evaluation of this study with the codebase [15].

The training data generation follows straightforward steps and includes common augmentation techniques. An instrumented client application accepts a "recipe" (a complete set of parameters, including those not considered as target variables here, see Figure 1) for constructing a character as an input. The parameters for the head shape get randomized within normalized limits imposed by the application. Next, the application renders and saves several views of the generated character while augmenting the imagery. The augmentation includes varying background, camera, lighting, pose, facial expression, and gaze direction. In this paper, we consider as target variables facial parameters corresponding to a neutral face; we exclude the view, lighting, pose, and expression parameters. The normalized continuous parameters of interest (blendshape weights in our case) map to the floating-point target vector. The discrete components use one-hot encoding occupying target vector slices of the size corresponding to the number of the options.

A direct approach to training a model mapping an image to the heterogeneous target is to train a CNN with a multi-part loss function $L$:

$$L = \sum_{i=1}^{N} v_i R_i + \sum_{i=1}^{N} w_i C_i \qquad (1)$$

Here, $N$ is the number of facial regions (seven in our main case study, and we keep only three for Makehuman experimentation). $R_i$ is either mean L2 or L1 loss for modifiers; $C_i$ is the cross-entropy loss for the discrete elements, and $v_i$, and $w_i$ weights that can be adaptive [36]. We use transfer learning with *inception3* [44] and *squeezenet* [37,48] from Pytorch [42] models zoo as the starting point. Transfer learning provides a reasonable accuracy relatively early in the training process, but complete training may take longer with ~10k input images randomized across all target dimensions. The longer training times in the direct approach are a disadvantage, slowing iterations on the models for the evolving customer product. Also, a monolithic model makes it hard to address inaccuracies in concrete facial regions. Finally, the amount of training data may be insufficient for reliable generalization due to the high dimensionality of the target space, potentially leading to various biases and overfitting. Training a model for a complete set of parameters (including expression, pose, camera, lighting, etc) as the target vector may follow the same supervised approach and be even more computationally intensive. These considerations motivate our decomposition approach.

## 4. ENSEMBLE ENGINEERING VIA DOMAIN DECOMPOSITION

The natural subdivision of a human face into regions like the nose, mouth, and alike leads to the target space's corresponding decomposition. Such decomposition is present in many parametric-based modeling applications, including Makehuman (see Figure 1). Equation (1), grouped by region, gives loss functions for local models with the following caveats.

One of the caveats is overcompleteness. An overcomplete parametric model may generate the same visual appearance with different parameter values. One common source of such over-completeness is head scale, when we reconstruct the head independently from the rest of the body. Fixing a particular scale variable (e.g., setting the head width to 1) in the training data normalizes parameters and eliminates over-completeness in our experiments.

The other issue to consider is related to the grouping of the parameters. We construct terms for group loss using parameters' names indicating their influence region. However, a group loss produced directly from naming conventions and corresponding to a specific facial area may include local (e.g., angle) and global features (e.g., placement of the facial feature relative to the other facial features). Manual engineering of the group loss to account for such subdivision could be error-prone and subject to frequent revisions in evolving client software.

We automatically address local-vs-global ambiguity by introducing learnable weights into the ensemble. Global features learned within local groups would result in predictions equal to an average value over the dataset. When learned as part of the aggregate complete model, they will result in a much better prediction. Their learned combination with the introduced weights will automatically reflect their roles and predictive power. One way to formulate training for the ensemble weights is to frame it as an optimization problem

$$E(w) = |\sum_{i=1}^{n_k} w_i^{(k)} M_i^{(k)}(D) - T(D)|^2 \to \min_w \qquad (2)$$

Here, $E(w)$ is the cumulative L2 error computed for weights $w^{(k)}_i$ corresponding to the group $k$ and target variable $i$. By running models $M^{(k)}$ on the training dataset $D$, we obtain predictions for each group $k$. Linear combination of the predictions with weights $w^{(k)}_i$ gives ensemble prediction. We compare it with the known target vector $T(D)$ and compute mismatch on the entire dataset as a function of $w$. The weights vectors $w^{(k)}_i$ dimensionality equals the dimensionality of the target parameters space. Also, we normalize the weights so that they sum to 1 for each target variable. That gives complementary weights $w_i$ and $1-w_i$ for the coordinates shared by local and global features.

Solving (2) for *w* is a straightforward coordinate-wise task. Not restricting weights to positive values allows for the utilization of consistent bias in either local or global models and can further improve the accuracy of the ensemble. Figure 2 summarizes our approach to constructing, training, and using the decomposition-based ensemble for inference.

A complete target space in our study has a dimensionality close to 400. Its combinatorial part has a complexity of $\sim 10^{11}$. Decomposition into subspaces radically reduces the dimensionality of the target spaces for each model we need to train. In our application, the largest modifiers subspace has a dimensionality of $\sim 20$. The largest number of discrete options for discrete elements for a region is under 100. Since the subspaces for discrete parts represent mutually exclusive choices, the combinatorial complexity reduces by many orders of magnitude. Combined with lower input resolution for local features, these factors allow for more efficient training with a smaller amount of training data necessary for sufficient sampling coverage of the target space.

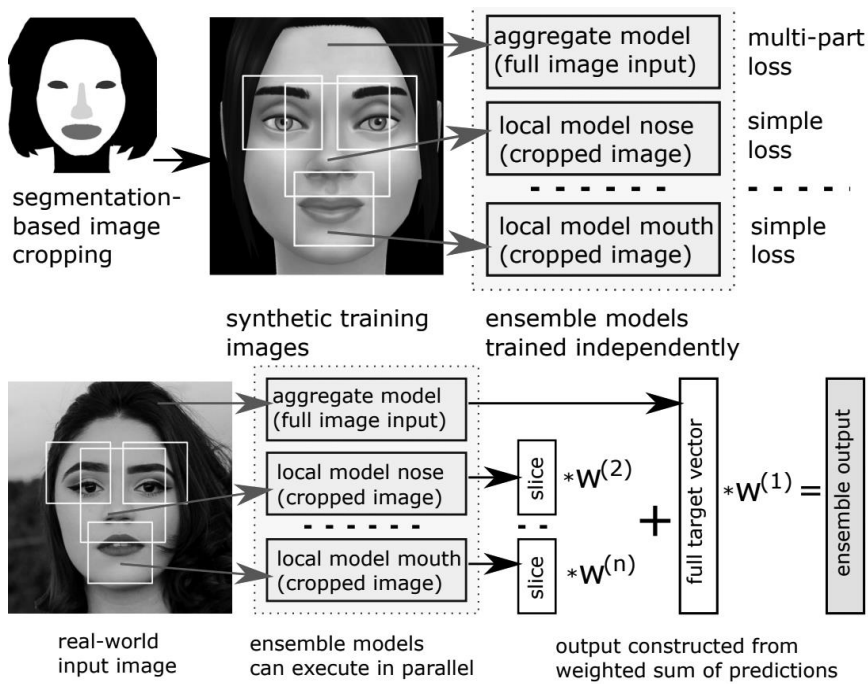

Figure 2. Training (top) and inference (bottom) with the decomposition-based ensemble. (We defer the domain adaptation part until the upcoming section, Figure 5.)

Since training models for the decomposed local features are less demanding than for an aggregate model, we may use smaller CNNs. In our experiments, we use the Pytorch model zoo's *squeezenet* for local features inference. Its smaller input resolution, 224x224, is suitable for cropped input image parts. Also, *squeezenet* is a lightweight model at only ~3Mb vs. *inception3*, which is close to 100Mb and takes 299x299 inputs. Our main study has 13 models, two per region (regression and classification grouped separately), with one region lacking discrete elements.

## 5. ENSEMBLE EVALUATION WITH A PROPRIETARY SOFTWARE

In our first round of experimentation, the proprietary software has a somewhat stylized, cartoonish rendering, defined by the art direction. Using automated objective evaluation (e.g., cosine distance in the latent space of FaceNet [43]) while disregarding the artistic style is problematic due to the domain gap. All our training and evaluation datasets are synthetic, rendered in the same artistic style, while we expect input images to come from photographs.

A panel-of-experts method provides a partial workaround to the limitations of the FaceNet evaluation. In our experiment, the panel includes experts evaluating the reconstruction quality of a hand-picked set of unannotated 20 facial images. The selection of the photos aims to represent various ages, ethnicities, lighting, image quality, poses, and facial expressions. The panel consists of nine professional artists and technical staff members. In the assessment, the respondents, besides other things, must rate each of the seven major facial regions of the reconstructed characters with a binary good/bad evaluation. Such simplified evaluation could be less informative than a popular Likert scale but makes the questionary easier and faster to complete. The results summary is in Table 1 and Figure 3.

Table 1. Comparison of ensemble vs. aggregate models by a panel of experts. Models' decomposition and ensembling reduce reconstruction defects by 30-50%. The worst possible defects score is 180 for nine experts and 20 images. The FaceNet cosine distance to the input image also reduces on most images by ~ 10%.

| Reconstruction defects by region | Cheeks | Chin | Eyes | Forehead | Jaw | Mouth | Nose |
|---|---|---|---|---|---|---|---|
| Naive aggregate model | 19 | 18 | 69 | 3 | 29 | 37 | 34 |
| Ensemble with models' decomposition | 6 | 11 | 34 | 1 | 16 | 24 | 17 |

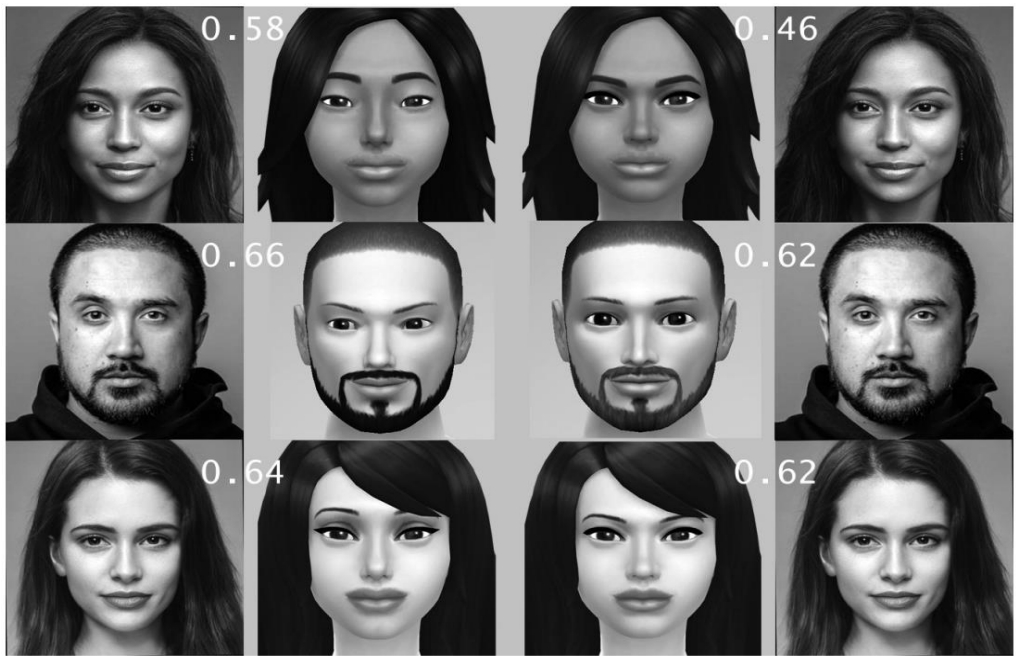

Figure 3. Naive aggregate model reconstruction (second on the left column) compared to the decomposition with the model's ensemble (second on the right). The numbers indicate FaceNet [43] cosine distance between the corresponding pairs. The imagery does not intend to represent any current or future commercial product. Photography attribution here (side columns) and in the following images [35].

While useful for episodical evaluation, the panel of experts' method is still time-consuming and hard to reproduce. Also, it isn't easy to maintain the same experts between evaluation rounds. To safeguard against such issues, we also computed the relative change in the FaceNet distance between input and reconstruction while being aware of the domain gap. The calculated difference is in sync with the experts' evaluation shown in Table 1. The table summarizes only the experts' responses who participated in both assessment rounds: one for the naive aggregate model and the other for the models' ensemble. The table highlights improvements in the selected quality metrics obtained by introducing the model ensemble and clearly shows the advantages gained from the decomposition approach.

# 6. DOMAIN GAP AND DOMAIN ADAPTATION

We can't ensure that synthetic training imagery covers the entire domain of the expected inputs or is its representative sub-domain. Also, in applications, we may encounter new domains not anticipated during training. The domain gap between synthetic and target imagery makes the models underperform.

Two major factors are leading to the domain gap. One is the limitations of the parametric model itself, which may not be powerful enough to generate a sufficient variety of faces. The second factor is the artistic stylization produced by the client application. The stylization lends itself to various domain adaptation techniques. Using style adaptation (e.g., see [14]) as an intermediate step between the input and the rest of the inference pipeline should improve accuracy loss. Namely, we propose using an "inverse" style adaptation from the input to the stylized synthetic imagery. We can train such an adapter for various inputs besides real-world images, e.g., fine art, anime, and sketches (see Figure 4).

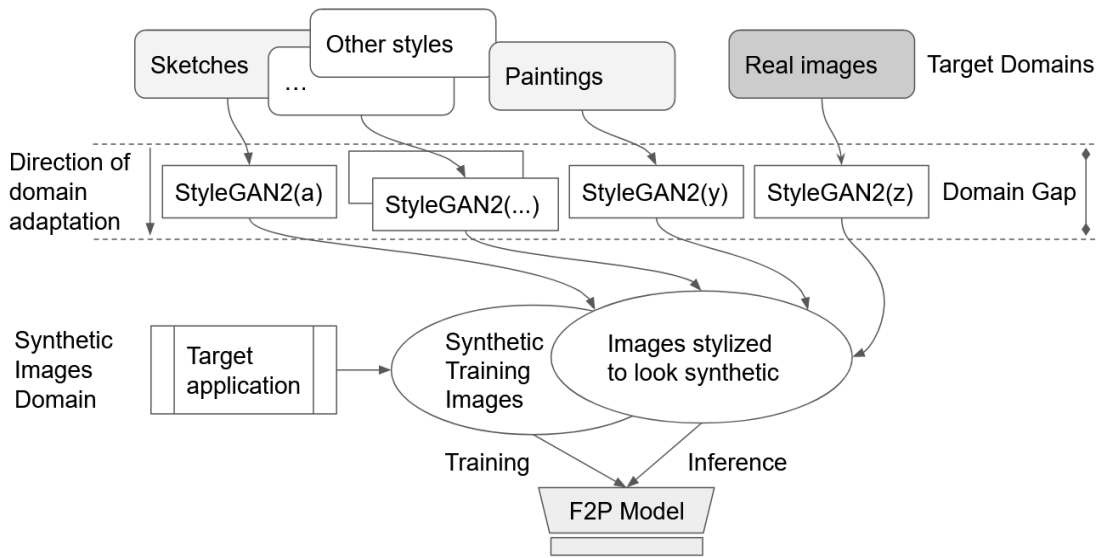

Figure 4. The F2P model uses for training synthetic images generated by an instrumented target application. When applied to real-world images, the domain adaptation with StyleGAN2 or a similar technique improves the accuracy of prediction (top-down arrows on the right side of the diagram). Similarly, we can train adapters for other target domains, e.g., sketches, paintings, etc. Note that stylized and synthetic image domains do not necessarily coincide.

After assessing various style transfer options (see Previous Works section), we use StyleGAN2 [34] for the academic part of our study (it has no permissive commercial license) to train and evaluate the proposed style adapter in application to our F2P inference pipeline. The synthetic training domain in our experiments comes from Makehuman [8], picked for its architectural similarity with our proprietary software. The target domain is real-world imagery. For inference, we apply StyleGAN2 to the input image to make it look like it comes from Makehuman, then feed it into the pipeline trained with Makehuman synthetic dataset. To train the adapter, we start with a StyleGAN2 pretrained on the FFHQ dataset (created for [40]) and fine-tune it on 4000 Makehuman-generated images. We normalize the images by registering (resizing and aligning) them using dlib landmarks [41] to match the images' alignment in the FFHQ dataset. After computing the style weights, the inference continues as follows. We start with a normalized real image. Next, we compute the latent projection vector for the given image through the StyleGAN2 mapping network and then apply the blended style weights from the selected resolution to map the real image to the stylized image. Figure 5 illustrates the results of that process.

Finally, we feed the stylized image to the inference F2P pipeline, which becomes an ensemble, including the domain adaptation step. The proposed approach applies to different input domains as in Figure 5 while keeping the same F2P CNN model.

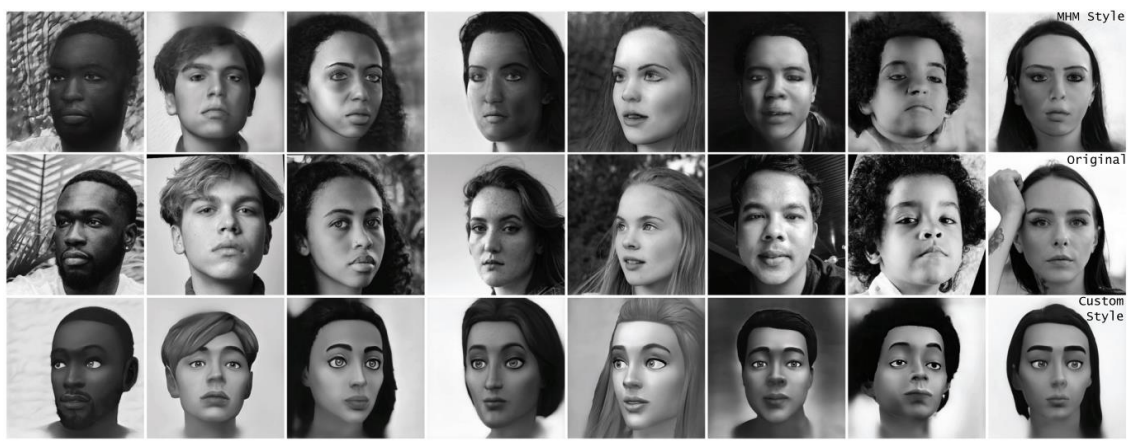

Figure 5. Stylized images from unsplash.com to the proprietary and Makehuman styles. Top row: Makehuman style, middle: original photos, Bottom row: proprietary style.

## 7. QUANTITATIVE METRICS AND ABLATION STUDY

We conduct a series of experiments with Makehuman software, reproducing the setup of our proprietary-based experimentation to support our findings using open-sourced, reproducible, quantifiable methods. For simplicity, we use a trivial baseline for the collected metrics: a mean over the evaluation dataset for the target variables. In our case, it corresponds to zero for all target variables describing an average face shape. Here, we focus on the regression part of the problem since Makehuman does not offer many discrete elements for the classification part. Besides, the decomposition advantage for classifiers follows directly from the combinatorial considerations.

Note that our goal for this concrete open-source study is not to train the best CNNs. Instead, we fix the training setup (meta-parameters, adaptive multi-part loss tuning, and the training schedule) to a reasonable common one and compare results in terms of accuracy between the combinations of the binary factors we describe after stating the following two claims to verify:

- **A decomposition-based ensemble** improves inference accuracy over the monolithic model.
- **Style adaptation via inverse style transfer** as a preprocessing step improves the accuracy of a model or models trained with synthetic training data.

For the decomposition-based ensemble claim, the goal is to obtain metrics characterizing the accuracy of the models trained either as a monolithic or decomposition-based ensemble. We evaluate the models utilizing the pretrained *squeezenet*, which we update in feature extraction and fine-tuning transfer learning modes. The motivation for considering both comes shortly. Overall, these three binary factors influence the results:

1. **Target space partitioning:** a complete target vector or limited to a particular feature group (e.g., nose parameters only). Such partitioning corresponds to the grouping terms of the target variables by features in multi-part loss (1).
2. **Input image:** either a complete frame or corresponding to a feature group "semantic" crop. We resize the cropped input image to match the CNN input.
3. **Transfer learning mode:** feature extraction or fine-tuning. In our tests, the fine-tuning phase starts from the CNN obtained by the features extraction step to speed up training.

Figure 6 illustrates the setup for binary factors we test, and Table 2 summarizes the results.

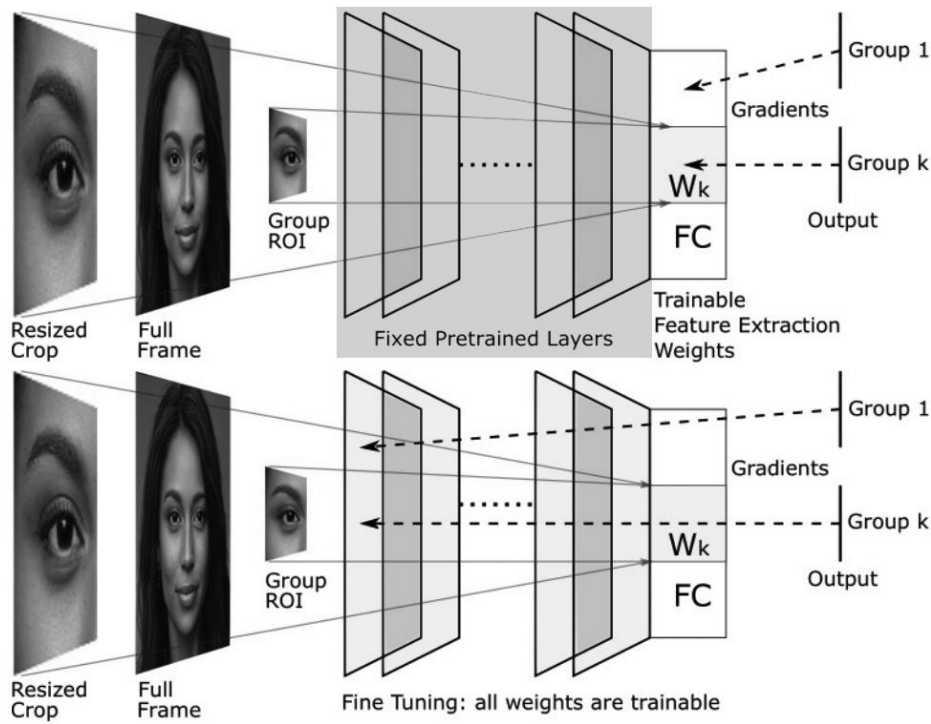

Figure 6. Quantitative evaluation setups and ablation study. We compare pairs: full frame vs. crop input and complete multi-part loss vs. local group loss. Top: transfer learning with feature extraction. Bottom: transfer learning with fine-tuning.

We start with the target space partitioning. One of the factors that may influence the models' training and the resulting accuracy is the difference in scale of the gradients in the multi-part loss (1). If such a difference is significant, it may take longer to achieve the required accuracy across all target variables. We train the first set of models using feature extraction transfer learning to isolate that factor. In that form, the CNN's hidden layers' weights remain fixed, with only the top fully connected (FC) layer trained to fit the dataset. The input in this experiment is a full-frame image for all models.

When training stops with a predefined learning rate schedule by reaching a plateau for the evaluation dataset, we expect similar results between the local and monolithic models compared to the local target features. Running training sufficiently long with a shared learning rate scheduler should reduce the different gradient scales' effect (regardless of the overfitting concerns). The top part of Figure 5 illustrates the proposed setup. The hidden layers inside the greyed box remain fixed. The grey FC weights $W_k$ for the target group $k$ train (nearly) independently in monolithic and local features variants. The main expected difference is the loss improvement rate but similar resulting accuracy. Hence, the decomposition-based ensemble should not provide a notable advantage over the monolithic model with feature extraction training with a full-frame input. Table 2 shows only a tiny improvement from complete to partial loss function when using the full frame as an input, supporting intuition.

Table 2. Local feature group models demonstrate the advantages of the hierarchical decomposition of the F2P problem. The inaccuracy (mean L1- loss) generally decreases left-right top-down within each group. The inaccuracy shown is relative to the baseline. Limiting the number of terms in multi-part loss and cropping the feature improves the resulting model by order of magnitude for some features. The resulting models' ensemble is far superior to the aggregate model trained on the complete input image while sharing with the aggregate one a similar training setup.

| Features Group | Loss | Input | Inaccuracy vs. baseline (smaller is better) | |
|---|---|---|---|---|
| | | | Feature Extraction | Fine Tuning |
| Nose | Complete | Full frame | 0.0001 | -0.0039 |
| | Local | Full frame | -0.0005 | 0.0007 |
| | | Cropped | -0.0059 | **-0.0743** |
| Mouth | Complete | Full frame | -0.0001 | -0.0135 |
| | Local | Full frame | 0.0005 | 0.0000 |
| | | Cropped | -0.0080 | **-0.0744** |
| Eyes | Complete | Full frame | -0.0003 | -0.0235 |
| | Local | Full frame | -0.0001 | 0.0010 |
| | | Cropped | -0.0061 | **-0.0530** |

Table 3. Fitting weights for the features individually reduces the prediction error compared to the simple ensemble with constant weights across all the dimensions. We set the baseline for this table per training type as the corresponding accuracy (validation loss) computed with constant weights (i.e., 0.0, 0.5, and 1.0). Smaller is better.

| Constant weights | Feature Extraction | | Fine Tuning | |
|---|---|---|---|---|
| | Full Frame | Crop | Full Frame | Crop |
| 0.0 (aggregate model only) | -0.08 | -0.09 | -0.11 | -0.12 |
| 0.5 (equal mix of aggregate and local models) | -0.08 | -0.11 | -0.23 | -0.27 |
| 1.0 (local models only) | -0.08 | -0.19 | -0.44 | -0.52 |

The other factor influencing the accuracy of the models is choosing the input image used to train the local feature CNN. We expect the results to improve by moving from a complete full frame to crops specific to the particular feature groups. That should work even in the feature-extraction transfer learning case. The Feature Extraction column in Table 2 confirms such an expectation. E.g., for the Nose features group, compare numbers in the feature extraction column between Full-Frame and Crop rows.

The progression from complete to local loss while using full-frame input does not improve accuracy much compared to the local loss and cropped images. Moving from feature extraction to fine-tuning, we adjust all weights in the CNN; see the bottom row of Figure 6. That makes learning the target features less constrained and provides a significant boost to the models' accuracy trained with local loss function corresponding to the feature image crop.

Moving from the full frame to the cropped image input provides the most improvement for fine-tuning. The bold numbers in Table 2 correspond to the proposed local models used for constructing the decomposition-based ensemble. They show the best accuracy across the evaluated combinations of the binary factors.

Finally, Table 3 shows that an ensemble with learned weights for its individual parameters outperforms its components or a weighted sum with fixed equal weights for the predictions.

This section would be incomplete without a discussion of domain adaptation. The inverse style transfer pre-processing step with StyleGAN2 trained on Makehuman images improves the ensemble's accuracy. We could not utilize validation loss for that evaluation. Instead, we use cosine distance on FaceNet embeddings [28]. The absolute distances shown in Table 4 may look substandard than the commonly accepted threshold of 0.51 for person identification. However, their values are meaningful only in relative terms and illustrate that the domain adaptation moves the input distribution in the right direction, beneficial for the model ensemble. The bottom row distances for full-frame and crop cases are smaller for the style-adjusted images than for the original ones.

The presented experiments support our claims that in addition to being more manageable, and easier to train, the ensemble-based approach takes advantage of the underlying properties of the models we use. It helps to train a better accuracy inference pipeline more practically.

Table 4. FaceNet cosine distance to the original input image shows the advantages of the style transfer as the domain adaptation step. Here we use a subset of CelebA [11].

| FaceNet cosine distance | Input | Feature Extraction | | Fine Tuning | |
|---|---|---|---|---|---|
| | | Full Frame | Crop | Full Frame | Crop |
| Original | [0.0] | $0.889_{\pm0.13}$ | $0.893_{\pm0.12}$ | $0.917_{\pm0.12}$ | $0.875_{\pm0.12}$ |
| Stylized | $0.495_{\pm0.11}$ | $0.889_{\pm0.10}$ | $0.895_{\pm0.11}$ | $0.892_{\pm0.11}$ | $0.868_{\pm0.12}$ |

## 8. DISCUSSION AND FUTURE WORK

The domain adaptation in our approach is a straightforward stylization of the input images used as input to the F2P CNN model. Training style transfer for different domains allows for reusing F2P CNN without lengthy re-training. However, if the expected domain can be fixed and well understood, the stylization of the training images may also deliver advantages as in [51]. In that work, artificial degradation of the images to simulate a security camera feed improved the accuracy of the face-recognition models in the wild. Similarly, style adaptation, i.e., from cartoonish to realistic, for training images is a promising area of exploration. Figure 7 illustrates the conceptual difference between the two ways of applying domain adaption. As an early experiment (see Figure 8), we converted a stylized synthetic texture computed by our proprietary software into a more realistic one. Next, we map it on the synthetic parametric model, as shown in Figure 6. The right column represents updated training imagery that delivers better results, a subject for future publications. While such tests show promise, the approach could be limited in practical applications due to the wide range of possible input images from an unexpected domain. In our internal test run of the inference pipeline, we observed sketches, portraits of anthropomorphic cats, anime, robots, and more. Fixing the input domain in such applications can make the pipeline too rigid and less performing.

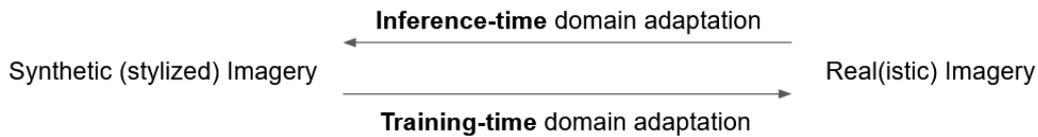

Figure 7. Training-time and inference-time domain adaptation. In this paper, we use inference time domain adaption ("inverse" style transfer, upper arrow in this Figure) with a StyleGAN2 model mapping real-world images to a synthetic stylized look. For a fixed, well-understood domain, an opposite approach may work too ("direct" style transfer, bottom arrow). In that approach, the entire training data set is pre-processed to deliver a look consistent with the target domain.

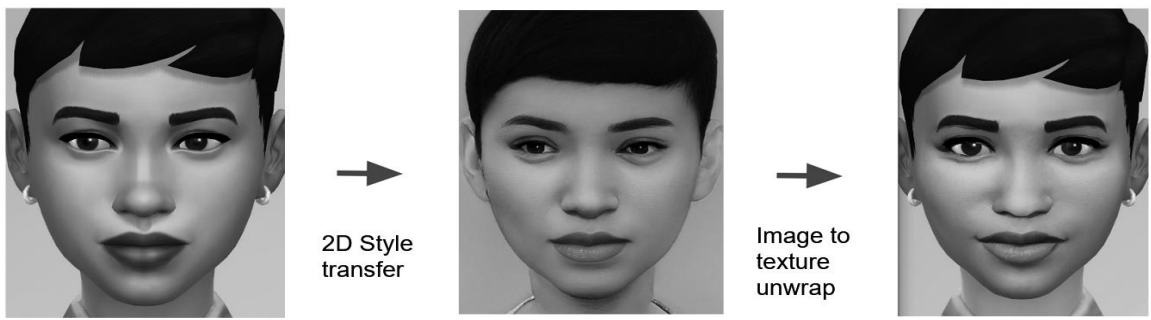

Figure 8. Bridging domains of synthetic and realistic textures. The input on the left is a synthetic image rendered with proprietary software. The middle image is the result of 2D style transfer with StyleGAN2. The original parametric 3D model textured with the unwrapped "realistic" texture is on the right. While the domain gap may persist due to limited lighting and rendering capabilities of the target software, the image on the right subjectively appears less synthetic.

The proposed in this paper technique is modular by design. Modularity simplifies maintenance and re-training components as needed in the production and application. A trade-off is additional moving parts. It may look tempting to settle for a single yet powerful enough monolithic model. However, combinatorial estimates for the discrete elements inference suggest that the amount of training data would still be exponentially larger, as discussed in Section 1.2. Such observations indicate that the modular decomposition design remains attractive and can be generalized further.

One way to generalize the proposed decomposition schema is to include aggregate spaces of parameters. An aggregate dimension would abstract several related parameters that tend to change correlatedly and wrap them into a single one. An obvious example is a facial expression like a smile controlled at a high level by one slider. Aggregating parameters defining a smile into a single parameter reflects a correlated change in the outline of the mouth, eyes, folds around the nose, etc. Returning to a neutral face reconstruction, we may express similar correlations regarding age, gender, ethnicity, and more. Construction of such aggregation may start with introducing a hierarchy to the blendshapes space representing common gross changes corresponding to the larger-scale features. The blendshapes hierarchy would naturally map to the partitioning and aggregation of the parametric space. The introduction of the Level of Details (LOD) commonly used for CG models in video games could also come naturally from such a hierarchy. A mobile application may use fewer aggregate blendshapes (lower LOD) instead of the complete detailed set of the blendshapes available on more capable platforms (high LOD). That line of thinking is inspired by the general decomposition methodology [54]. The principled approach to complex systems explained in [51] demonstrates its effectiveness in many applications. However, to benefit from such an approach, the area of facial likeness reconstruction may require additional experimentation and a deeper understanding of the anatomical and visual structure of the human face in terms of modeling parameters. A Bayesian network may be a good approach to building a composite model that integrates facial features across scales, aggregations, and locality. Such integration would address our primary assumption of the target features' independence which is an oversimplification of a correlated set of facial features. Including such correlations in the proposed ensemble may further improve the accuracy and consistency of the predictions. The Bayesian network would replace a simple weighted sum with belief propagation and integrate otherwise ignored correlations between local features. It appears to be a valid subject to explore in the future. Since decomposition is a general methodology, other parametric models besides human faces may benefit from the proposed approach. The proposed network of models would remain modular and the belief propagation can be implemented as lightweight post-processing.

So far, we have discussed monocular reconstruction from RGB images. A depth channel on some input devices suggests an attractive avenue of exploration that includes depth into the model input, e.g., [55]. The depth channel may still have limited precision, and the main power of monocular reconstruction would still rely on the RGB image. Another venue of generalization is using video as input. It may appear straightforward (e.g., [45,46]), but proper averaging models with confidence or accuracy weights may require additional work.

For completeness, we mention here the limitations of the proposed technique. Reconstruction of a human face in the F2P problem is a multi-faceted task. The paper covers only a single subject – the decomposition of "geometric" features that include continuous and discrete elements. In our experience, the classification of glasses, earrings, and some other localized features also benefit from the segmentation and decomposition approach. However, many "spread" features (e.g., hairstyle) do not easily lend themselves to the proposed method and are a subject for future exploration.

## 9. CONCLUSION

The paper proposes a novel combination of well-established domain manipulation techniques summarized in Figure 9.

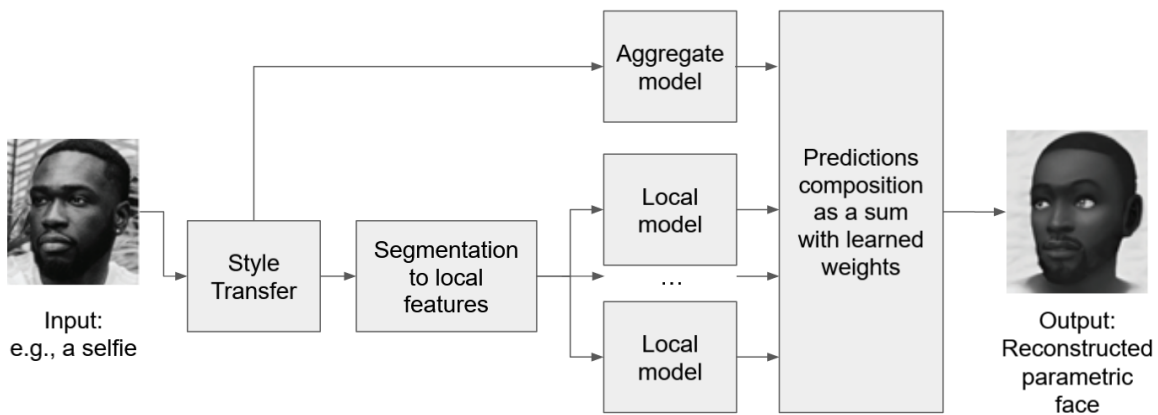

Figure 9. The diagram summarizes the proposed inference pipeline based on domain adaptation and decomposition ensemble. Style transfer, like the local models, is a pluggable module with the corresponding model easily re-trainable as the requirements change.

Despite its conceptual simplicity, the domain decomposition combined with domain adaptation provides several measurable benefits in the F2P problem that are particularly valuable in the applications to external user-facing software and the in-house art production pipelines. It facilitates training of the models, offers better control over their accuracy, convenient maintenance vital for industrial applications, smaller memory footprint during inference, and flexibility across input domains. The proposed approach is not fundamentally limited to parametric faces and may work for similar problems in other computer vision applications.

## AUTHORS

**Igor Borovikov** received his Ph.D. in math at Moscow Institute for Physics and Technology in 1989. He is a Senior AI Scientist at Electronic Arts.

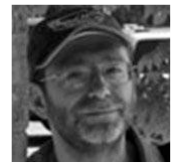

**Karine Levonyan** received her Ph.D. at Stanford University in 2019. She is an AI Scientist III at Electronic Arts.

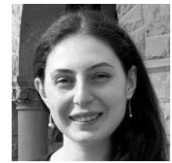

**Jon Rein** graduated from Art Institute (CDIS) in 2004. Currently, he is a Senior Software Engineer at Electronic Arts.

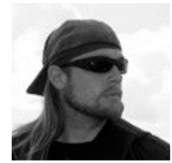

**Pawel Wrotek** graduated from Brown University in 2006 and received M.S. in Computer Science. He is currently a Senior Software Engineer at Electronic Arts.

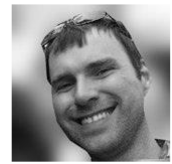

**Nitish Victor** received his M.S. from Rochester Institute of Technology in Game Design and Development in 2019. He is currently a Software Engineer II at Electronic Arts.

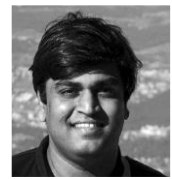